\documentclass[10pt,twocolumn,letterpaper]{article}

\usepackage[pagenumbers]{cvpr} 

\usepackage{graphicx}
\usepackage{amsmath,bm}
\usepackage{amssymb}
\usepackage{booktabs}
\usepackage{makecell}
\usepackage{multirow}
\usepackage{cuted}
\usepackage{capt-of}

\usepackage[pagebackref,breaklinks,colorlinks]{hyperref}

\usepackage[capitalize]{cleveref}
\crefname{section}{Sec.}{Secs.}
\Crefname{section}{Section}{Sections}
\Crefname{table}{Table}{Tables}
\crefname{table}{Tab.}{Tabs.}

\def\cvprPaperID{1805} 
\def\confName{CVPR}
\def\confYear{2023}

\begin{document}

\title{High-fidelity Facial Avatar Reconstruction from Monocular Video with Generative Priors}


\author{%
  Yunpeng Bai$^{1}$, Yanbo Fan$^{2}$, Xuan Wang$^{3}$, Yong Zhang$^{2}$, Jingxiang Sun$^{1}$,  Chun Yuan$^{1,4}$, Ying Shan$^{2}$ \\[0.5em]
  $^{1}$ Tsinghua University, $^{2}$Tencent, $^{3}$Ant Group, $^{4}$Peng Cheng Laboratory \\[0.3em]
}


\maketitle

\begin{strip}
\centering
\includegraphics[width=\textwidth]{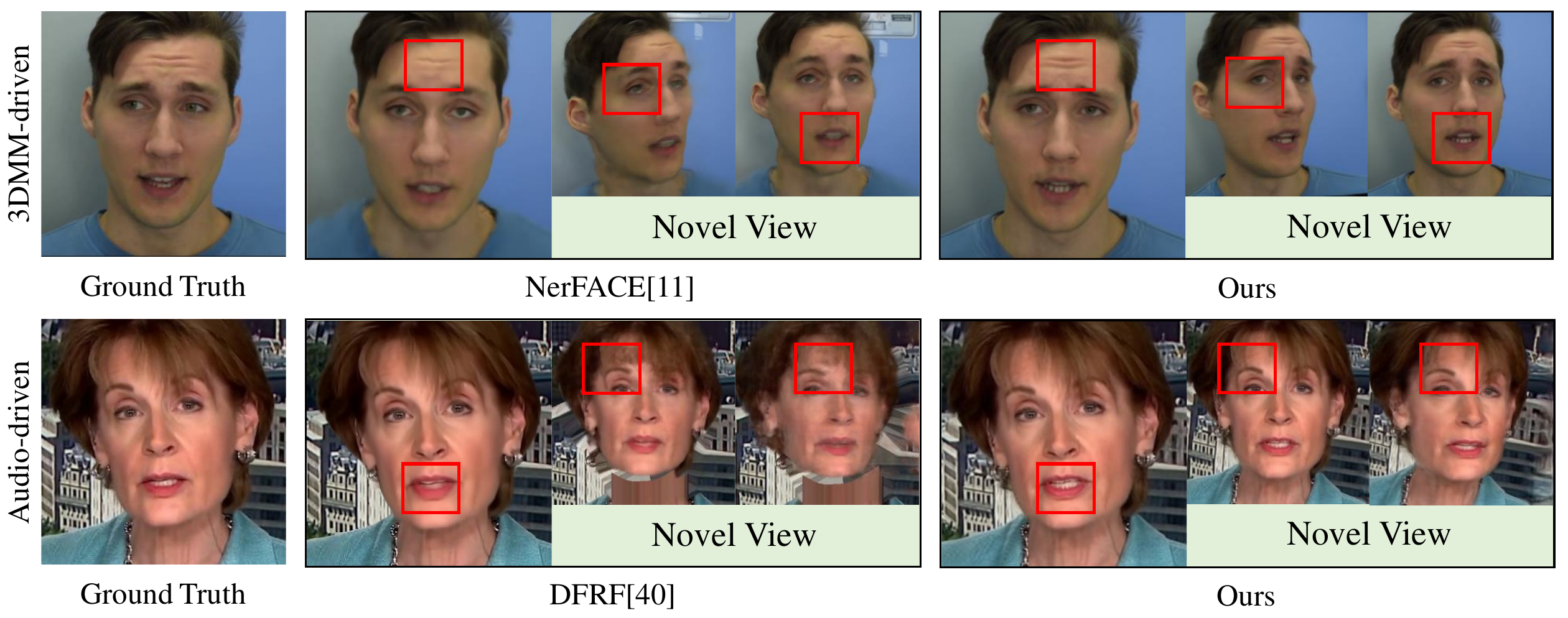}
\captionof{figure}{Visualizations of 3DMM and audio-driven face reenactment of our proposed method and NerFACE~\cite{gafni2021dynamic} and DFRF~\cite{shen2022learning}. 
The left-most column is the ground truth images.
For each method, the left plot is the rendered image with the same view of the ground truth image, and the two right plots are novel view syntheses.
By utilizing the high-quality 3D-aware generative prior, our method significantly boosts the performance of face reenactment and novel view synthesis. We highlight some areas with red rectangles for better comparisons.
}
\label{fig:teaser}
\end{strip}


\begin{abstract}


High-fidelity facial avatar reconstruction from a monocular video is a significant research problem in computer graphics and computer vision.
Recently, Neural Radiance Field (NeRF) has shown impressive novel view rendering results and has been considered for facial avatar reconstruction.
However, the complex facial dynamics and missing 3D information in monocular videos raise significant challenges for faithful facial reconstruction.
In this work, we propose a new method for NeRF-based facial avatar reconstruction that utilizes 3D-aware generative prior.
Different from existing works that depend on a conditional deformation field for dynamic modeling, we propose to learn a personalized generative prior, which is formulated as a local and low dimensional subspace in the latent space of 3D-GAN.
We propose an efficient method to construct the personalized generative prior based on a small set of facial images of a given individual.
After learning, it allows for photo-realistic rendering with novel views and the face reenactment can be realized by performing navigation in the latent space. 
Our proposed method is applicable for different driven signals, including RGB images, 3DMM coefficients, and audios. 
Compared with existing works, we obtain superior novel view synthesis results and faithfully face reenactment performance.

\end{abstract}

\section{Introduction}
\label{sec:intro}
Reconstructing high-fidelity controllable 3D faces from a monocular video is significant in computer graphics and computer vision and has great potential in digital human, video conferencing and AR/VR applications.
Yet it is very challenging, due to the complex facial dynamics and missing 3D information in monocular videos.




Recently, Neural Radiance Field (NeRF) \cite{mildenhall2021nerf} has shown impressive quality for novel view synthesis.
The key idea of NeRF is to encode color and density as a function of spatial location and viewing direction by a neural network and adopt volume rendering technique for novel view synthesis.
Its photo-realistic rendering ability has sparked great interest in facial avatar reconstruction.
Starting with static face reconstruction, deformable neural radiance field has been proposed to handle the non-rigidly deforming faces captured in monocular videos.
For example, the works of \cite{park2021nerfies,park2021hypernerf} proposed to learn a conditional deformation field to capture the non-rigidly deformation of each frame.
After training, they can provide novel view synthesis for the training frames.
However, they don't support facial editing and cannot be used for face reenactment. 

The controllability of facial avatar is indispensable for many down-stream applications, such as talking head synthesis.  
To facilitate this, the core idea of existing works is to learn a dynamic neural radiance field conditioned on specific driven signals.
For example, 3D morphable face model (3DMM) \cite{blanz1999morphable} is introduced as a guidance in NeRF-based facial avatar reconstruction \cite{gafni2021dynamic,athar2022rignerf,gao2022reconstructing}.
The work of \cite{gafni2021dynamic} learns a dynamic NeRF that directly conditioned on the pose and expression coefficients estimated by 3DMM.
In RigNeRF \cite{athar2022rignerf}, the deformation field is a combination of a pre-calculated 3DMM deformation field prior and a learned residual conditioned on the pose and expression coefficients.
After modeling, one can use 3DMM coefficients for face reenactment.
In addition to the explicit 3DMM coefficients, audio-driven dynamic NeRF has also been studied \cite{guo2021ad,shen2022learning}.
Recently, AD-NeRF \cite{guo2021ad} is proposed to optimize a dynamic neural radiance field by augmenting the input with audio features.
DFRF \cite{shen2022learning} further considers the few-shot audio-driven talking head synthesis scenario.
%
These works directly learn a conditional deformation field and scene representation in the continuous 5D space.
However, recovering 3D information from monocular videos is an ill-posed problem, it is very challenging to obtain high-fidelity facial avatar.

To alleviate the aforementioned challenges, we propose to adopt 3D generative prior.
Recently, 3D-aware generative adversarial networks (3D-GAN) \cite{chan2021pi,chan2022efficient,gu2021stylenerf,or2022stylesdf,sun2022ide} are proposed for unsupervised generation of 3D scenes.
By leveraging the state-of-the-art 2D CNN generator \cite{karras2020analyzing} and neural volume rendering, the work of \cite{chan2022efficient} can generate high-quality multi-view-consistent images.
The latent space of 3D-GAN constitutes a rich 3D-aware generative prior, which motivates us to explore latent space inversion and navigation for 3D facial avatar reconstruction from monocular videos.
However, 3D-GAN is usually trained on the dataset with a large number of identities, such as FFHQ \cite{karras2019style}, resulting in a generic generative prior.
It is inefficient for personalized facial reconstruction and reenactment, which requires faithfully maintenance of personalized characteristics.

In this work, we propose to learn a personalized 3D-aware generative prior to reconstruct multi-view-consistent facial images of that individual faithfully.
Considering that facial variations share common characteristics, we learn a local and low-dimensional personalized subspace in the latent space of 3D-GAN.
Specifically, we assign $k$ learnable basis vectors $\mathbf{B} \in R^{k\times d}$ for the individual, where $d$ is the dimension of the 3D-GAN latent space.
Each frame is sent to an encoder to regress a weight coefficient $\bm{\alpha} \in \mathcal{R}^{1\times k}$, which is used to form a linear combination of the basis.
The resulting latent code is sent to a 3D-aware generator for multi-view-consistent rendering.
We show that both the personalized basis and encoder can be well modeled given a small set of personalized facial images.
After training, one can directly project the testing frames with different facial expressions onto the learned personalized latent space to obtain a high-quality 3D consistent reconstruction.  
It is worth noting that the input modality is not limited to RGB frames.
We demonstrate with a simple modification. The encoder can be trained with different signals, such as 3DMM expression coefficients or audio features, 
enabling 3DMM or audio-driven face reenactment.
To verify its effectiveness, we conduct experiments with different input modalities, including monocular RGB videos, 3DMM coefficients, and audios.
The comparison to state-of-the-art methods demonstrates our superior 3D consistent reconstruction and faithfully face reenactment performance.

Our main contributions are four-fold:
1) we propose to utilize 3D-aware generative prior for facial avatar reconstruction;
2) we propose an efficient method to learn a local and low-dimensional subspace to maintain personalized characteristics faithfully;
3) {we develop 3DMM and audio-driven face reenactment by latent space navigation;}
4) we conduct complementary experimental studies and obtain superior facial reconstruction and reenactment performance. 

\section{Related Work}
\label{sec:related-work}
We introduce recent works that are closely related to our method, including neural volume rendering, controllable face generation, and generative 3D-aware neural networks.

\paragraph{Neural scene representation and rendering.} 
Recently, Neural Radiance Field (NeRF) \cite{mildenhall2021nerf,chibane2021stereo,gafni2021dynamic,gao2020portrait,li2021neural,lombardi2021mixture,martin2021nerf,pumarola2021d,xian2021space,zhang2020nerf++,peng2021neural,sun2022fenerf,ma2022neural,chen2022hallucinated} obtains impressive performance for novel view synthesis of complex scenes.
Instead of explicitly modeling the geometry and appearance, NeRF represents a scene using a neural network (\eg, MLP) to encode color and density as a function of a continuous 5D coordinate (including spatial location and viewing direction).
It then use classic volume rendering techniques for novel view synthesis.
The volume rendering is differentiable and the neural representation can be optimized given a set of images of a scene.

The photo-realistic 3D consistent rendering ability of NeRF has sparked great interest in facial avatar reconstruction.
However, the standard formulation in \cite{mildenhall2021nerf} is proposed for static scene representation.
And it requires multi-view input images for faithfully reconstruction.
To handle the non-rigid dynamics in facial images captured by a monocular camera, the work of \cite{park2021nerfies} proposed to learn a continuous deformation field, which learns a per-frame latent code and maps each observation coordinate into a canonical template canonical coordinate space.
Furthermore, HyperNeRF \cite{park2021hypernerf} proposed to learn a higher-dimensional deformation field to better model the topology variations.
After learning, they can be used for novel view synthesis.
However, they don't support facial editing.

\paragraph{Controllable face generation.}
Controllable face generation is a key building-block for many applications in computer graphics and computer vision.
The explicit 3D Morphable Face Model (3DMM) \cite{blanz1999morphable,cao2013facewarehouse,li2017learning} uses linear subspace to control pose, expression and appearance independently. 
It provides fine-grained control over poses and expressions.
However, it only models the face region and lacks personalized attributes, including hairs, eyes and accessories such as glasses.
It suffers from artifacts when used for photo-realistic rendering.
Apart from the explicit 3D based models, there have been several works that directly animate images in 2D space for face reenactment \cite{siarohin2019first,ren2021pirenderer,nirkin2019fsgan,wang2022latent,kowalski2020config,yin2022styleheat,prajwal2020lip,suwajanakorn2017synthesizing,taylor2017deep,zhang2021flow,wang2021one,ji2021audio}.  
They are usually realized by learning a warping field from driven information (\eg, image or audio) or training an encoder-decoder-based image translation networks.
These methods, however, have to learn 3D deformation from 2D input.
They couldn't provide free view synthesis and suffer from artifacts, especially for large driven poses or expressions.

Recently, some works are proposed for controllable NeRF-based facial avatar reconstruction.
They are realized by optimizing a conditional deformation field and scene representation based on either 3DMM coefficients or audio signals.
For example, the work of \cite{gafni2021dynamic} first transforms the camera space point into canonical space by the estimated pose parameters and then regresses its color and density conditioned on the 3DMM expression coefficients.
In RigNeRF \cite{athar2022rignerf}, the deformation field is realized as a combination of a explicit 3DMM deformation field and a predicted residual.
The deformed point, as well as the 3DMM expression and pose coefficients, are send to an MLP to predict the color and density.
To enable semantic control over facial expression, the work of \cite{gao2022reconstructing} learns a series of neural radiance fields as basis, and associate them with expression coefficients extracted by mesh-based face models.
As for audio-driven facial avatar, AD-NeRF \cite{guo2021ad} augments the 5D input with an audio feature for neural scene representation.
DFRF \cite{shen2022learning} proposed an audio-driven few-shot talking head synthesis method.
It learns a dynamic NeRF conditions on both audio features and 2D appearance images.
These methods, however, directly construct dynamic facial neural radiance field from a monocular video.
Considering the non-rigidly facial dynamics and missing 3D information in monocular videos, it is challenging to obtain high-fidelity multi-view-consistent results.
Instead of directly learning the dynamic radiance field, we propose to utilize the rich generative prior of 3D-GAN and learn facial avatar by latent space inversion and navigation.

\paragraph{3D-aware Generative Neural Networks.} 
Generative adversarial networks have achieved great success in image generation.
While most existing works focus on 2D images \cite{goodfellow2020generative,karras2017progressive,karras2019style,karras2020analyzing,nitzan2022mystyle}, recently 3D-aware generation has attracted more and more attention \cite{chan2021pi,chan2022efficient,zhou2021cips,gu2021stylenerf,or2022stylesdf,sun2022ide}.
The representative method, EG3D \cite{chan2022efficient} significant improves the quality of 3D-aware generation.
EG3D inherits the high-fidelity 2D image generation ability of StyleGAN \cite{karras2020analyzing} and the multi-view-consistent geometry of neural volume rendering.
It is shown that after trained on FFHQ \cite{karras2019style}, a real world large-scale face dataset, EG3D obtains state-of-the-art 3D face synthesis.
And its latent space constitutes a generative prior for multi-view-consistent images and 3D geometry.
Inspired by these, we propose to learn a personalized 3D generative prior to faithfully reconstruct the specific characteristic of a given individual.

\section{Proposed Method}
\label{sec:method}



\begin{figure*}
\centering
\includegraphics[width=0.95\linewidth]{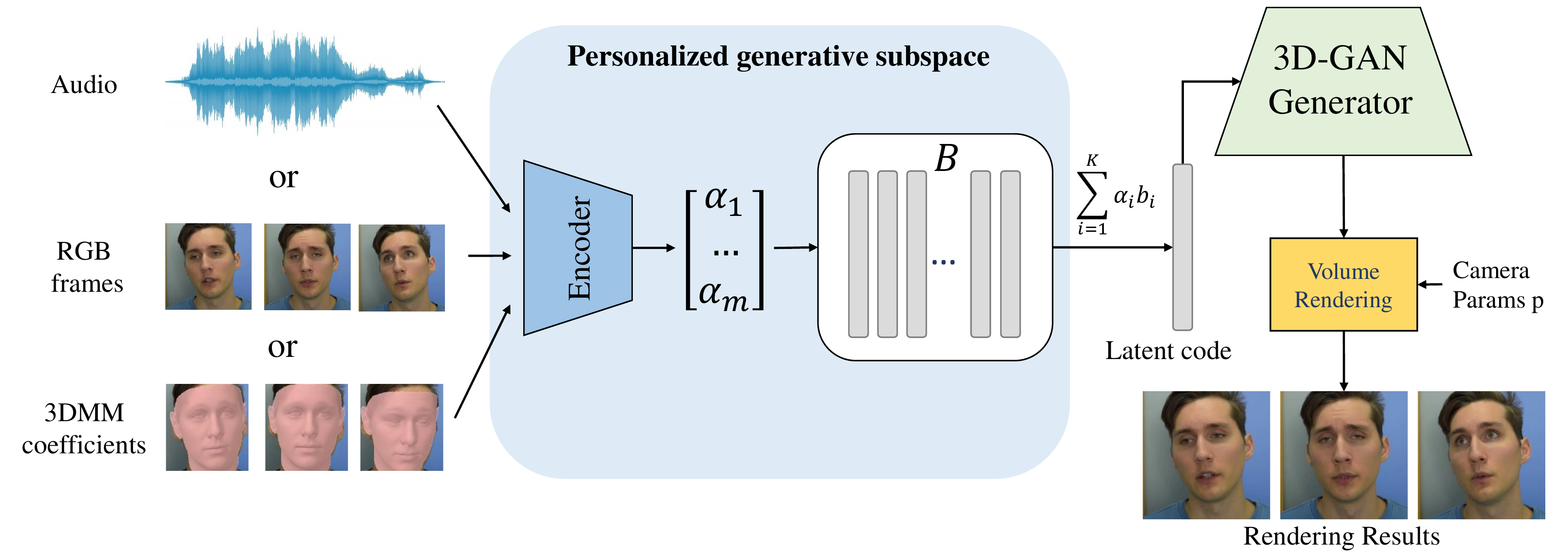}
\captionof{figure}{Overall framework of our proposed method. 
We assign a learnable personalized basis with $k$ vectors as $\mathbf{B}=[\mathbf{b}_{1}, \cdots, \mathbf{b}_{k}] \in \mathcal{R}^{k \times d}$ in the $\mathcal{W}+$ space. 
The input information (RGB frame, 3DMM expression coefficients, or audio features) is projected into the low-dimensional subspace $\mathcal{S}_{\mathbf{B}}$ by an encoder $\bm{f}$ as $\bm{w}=\bm{f}(\bm{x})\cdot \mathbf{B}$. 
$\bm{w}$ is then sent to 3D-GAN generator for free view synthesis.}
\label{fig:overall framework}
\end{figure*}

\subsection{Preliminary of 3D-GAN}
Our work builds on the multi-view-consistent image synthesis ability of 3D-GAN. 
The state-of-the-art method, EG3D \cite{chan2022efficient}, proposes an expressive hybrid explicit-implicit network based on a 2D CNN generator and neural rendering.
For a 3D-aware generation, each random sampled latent code is first sent to a pose-conditioned StyleGAN2 generator to learn a tri-plane 3D representation. 
It then learns a neural radiance field based on it and generates a low-dimensional raw image by volume rendering.
Finally, a super resolution module is adopted to generate high-resolution result.
After being trained on FFHQ \cite{karras2019style}, EG3D can be used for unsupervised generation of multi-view-consistent facial images. 
Its latent space constitutes a generative prior for facial images with consistent 3D geometry.

\textbf{A closer look at EG3D latent space.}
EG3D \cite{chan2022efficient} is an unconditioned generation network and doesn't provide any controls over identity or expression.
For generation w.r.t. to a specific facial image, one can invert the input image back into the latent space.
Given the inverted latent code, novel view synthesis can be realized by changing the camera pose during generation.
However, recovering the 3D geometry from a single image is an ill-posed problem.
Directly inverting facial images to the generic latent space of EG3D cannot faithfully reconstruct the specific characteristics of that individual, {an example is given in Figure~\ref{fig:pti}}.
In addition, it doesn't support face reenactment.

\subsection{Learning A Personalized Generative Prior}
We aim to reconstruct 3D-aware animatable facial avatar based on a monocular video.
To take advantage of the rich generative prior of 3D-GAN as well as maintain the personalized characteristics, we propose to learn a personalized generative prior.
Our overall framework is given in Figure~\ref{fig:overall framework}.
Specifically, we define the personalized generative prior as a local, low-dimensional, and smooth subspace in the latent space. 
The low-dimensional property is expected as the facial images of a specific identity share common properties.

We consider $\mathcal{W}+$ space of 3D-GAN and assign a learnable personalized basis with $k$ vectors as $\mathbf{B}=[\mathbf{b}_{1}, \cdots, \mathbf{b}_{k}] \in \mathcal{R}^{k \times d}$ in the $\mathcal{W}+$ space.
The subspace that spanned by $\mathbf{B}$ is defined by
\begin{equation} \textstyle
\mathcal{S}_{\mathbf{B}} = \left\{\bm{w} | \bm{w} = \sum_{i=1}^k \bm{\alpha}_i \bm{b}_i, \bm{\alpha} \in \mathcal{R}^{1 \times k}\right\},
\end{equation}
where $\bm{\alpha}$ represents the coefficient w.r.t. basis vectors.
Rather than directly inversing each facial image $\bm{x}$ into the high-dimensional $\mathcal{W}+$ space, we project $\bm{x}$ into the low-dimensional subspace $\mathcal{S}_{\mathbf{B}} $, by learning an encoder $\bm{f}: \bm{x} \rightarrow \mathcal{R}^{1 \times k}$ to regress the coefficient $\bm{\alpha}$ of $\bm{x}$.
Finally, $\bm{w}=\bm{f}(\bm{x})\cdot \mathbf{B}$ is sent to 3D-GAN generator for free view synthesis, as $\hat{\bm{x}}=\mathcal{G}(\bm{f}(\bm{x}), \mathbf{B}, \bm{p})$,
where $\mathcal{G}$ is the 3D-GAN generator and $\bm{p}$ is the camera pose used for rendering.

\subsection{Training Objective}
\label{sec:training objective}
Given a monocular face video $\bm{X} = \{\bm{X}^t\}^T_{t=1} \in \mathcal{R}^{T\times H\times W \times 3}$ of a individual with $T$ frames, each frame of which contains different expressions and poses.
The encoder $\bm{f}$ and the basis $\mathbf{B}$ are jointly optimized for a faithfully reconstruction of $\bm{X}$. 
Let $\hat{\bm{X}}^t = \mathcal{G}(\bm{f}(\bm{X}^t), \mathbf{B}, \bm{p}^t)$ and $\bm{p}^t$ is the camera pose extracted from $\bm{X}^t$, we calculate the $\mathcal{L}_2$ loss and LPIPS loss \cite{zhang2018unreasonable} between $\bm{X}^t$ and $\bm{X}^t$,
\begin{equation} \textstyle
    \mathcal{L} = \sum_{t=1}^{T} \mathcal{L}_2(\bm{X}^t, \hat{\bm{X}}^t) + \lambda_{lpips}\mathcal{L}_{lpips} (\bm{X}^t, \hat{\bm{X}}^t).
\end{equation}

During training, we further constraint the basis vectors to be orthogonal to each other.
The orthogonal constraint can largely boost the disentanglement of the basis.
{We provide an visualization of the basis in Figure~\ref{fig:ortho}.}
For the generator $\mathcal{G}$, we adopt the pretrain model \cite{chan2022efficient} that learned on FFHQ. 
Similar to PTI \cite{roich2022pivotal}, during training we slightly modify the generator to better maintain the personalized characteristics.
We use a two-stage training strategy: 1) freezing the parameters of the generator and updating the encoder $\bm{f}$ and the basis $\mathbf{B}$,
and 2) turning on the gradient of the generator to adapt it to the personalized subspace.

\textbf{Generalize to testing frames.}
The local and low-dimensional personalized subspace provides a good generalization to facial variations beyond the training frames.
After training, the encoder $\bm{f}$ can be directly applied to testing frames with different facial expressions to generate high-fidelity facial avatar reconstruction.

\subsection{Face Reenactment with Various Signals}
In the above section, we learn an encoder to project each facial image into the personalized latent space, to provide faithful 3D-aware reconstruction.
Indeed, the input signal is not limited to facial images.
Here, we provide two realizations with 3DMM coefficients and audio signals as input information, respectively.
After training, they can be used for 3DMM or audio driven 3D-aware face reenactment.

    \textbf{3DMM-driven face reenactment:} 
    we extract 3DMM expression coefficient $\bm{\beta} \in \mathcal{R}^{76}$ from each image, forming training pairs of $\{(\bm{X}^t, \bm{\beta}^t)\}_{t=1}^{T}$.
    The coefficient $\bm{\alpha}$ is learned by $\bm{\alpha}=\bm{f}_{e}(\bm{\beta})$.
    %
    \textbf{Audio-driven face reenactment:} 
    following \cite{guo2021ad}, we use {\it DeepSpeech} \cite{amodei2016deep} to extract a 29-dimensional feature for each 20ms audio clip. 
    To eliminate the noisy signals from raw input, we concatenate the features of sixteen neighboring audio clips, resulting in $\bm{\delta} \in \mathbb{R}^{16 \times 29}$ for the audio feature of the current frame.
    We then project $\bm{\delta}$ into the latent space by $\bm{\alpha} = \bm{f}_{a}(\bm{\delta})$. 

The realization of $\bm{f}_e$ and $\bm{f}_a$ are given in the {\it supplementary materials}.
We follow the training process in Sec.\ref{sec:training objective} for the learning of $\bm{f}_e$, $\bm{f}_a$ and their corresponding basis.
We compare to existing 3DMM and audio-driven face reenactment in Sec.\ref{sec:experiments-3dmm} and Sec.\ref{sec:audio-driven experiments}.
%


\section{Experiments}
\label{sec:experiments}
Our method performs 3D facial reconstruction with a monocular video sequence, and the modeled face can be driven by various input signals.
In this section, we first introduce the experimental settings. Then, we perform the reconstruction with RGB images, 3DMM coefficients, and audio signals as inputs, respectively. We also compared our proposed method with several baseline models both qualitatively and quantitatively. Finally, we performed several ablation studies to analyze the key elements of our approach.

\subsection{Implementation Details}

\textbf{Data preprocessing.} 
The training video needs to be processed before face modeling. 
For each frame of the video, we use an off-the-shelf pose estimator \cite{deng2019accurate} to estimate its corresponding camera intrinsic and extrinsic matrices as the input to EG3D generator. The flattened $4 \times 4$ camera extrinsic matrix and flattened $3 \times 3$ camera intrinsic matrix are concatenated into a 25-dimensional vector as the camera input to EG3D generator. Then, we extract the appropriately-sized crops from each frame and resize each cropped image to the resolution of $512 \times 512$.

\textbf{Training details.} 
For each video, we train the network 200k iterations to obtain a personalized face model. 
The parameters of the generator are not optimized in the first 50k rounds. We train our model on a single Nvidia Telsa V100 GPU. We use the Adam optimizer \cite{kingma2014adam} to train the network, and the learning rate is set to $3e^{-4}$, $\beta_1$ and $\beta_2$ set to $0.9$ and $0.999$, respectively. 
The batch size is $2$ and $\lambda_{lpips}= 5$.
The number of basis vectors in our method is set to $k=50$.




\begin{figure}[t]
    \centering
    \includegraphics[width=\linewidth]{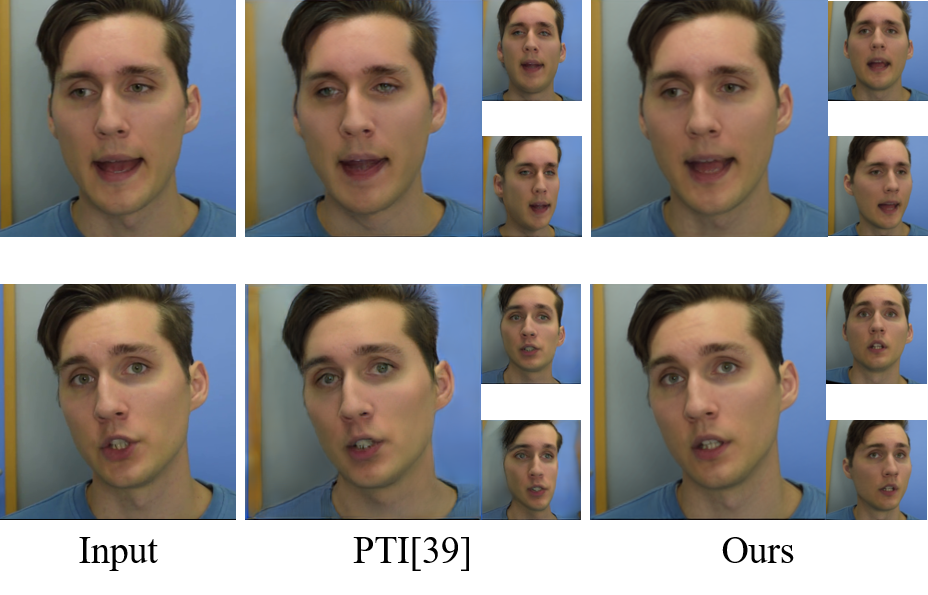}
    \caption{Visualizations of face reconstruction. The left-most column is the input image. For PTI and ours, we plot rendered images with the input camera view (the big plot in the left side) and two novel views (the two small images in the right side). }
    \label{fig:pti}   
\end{figure}

\begin{figure*}[t]
    \centering
    \includegraphics[width=0.95\linewidth]{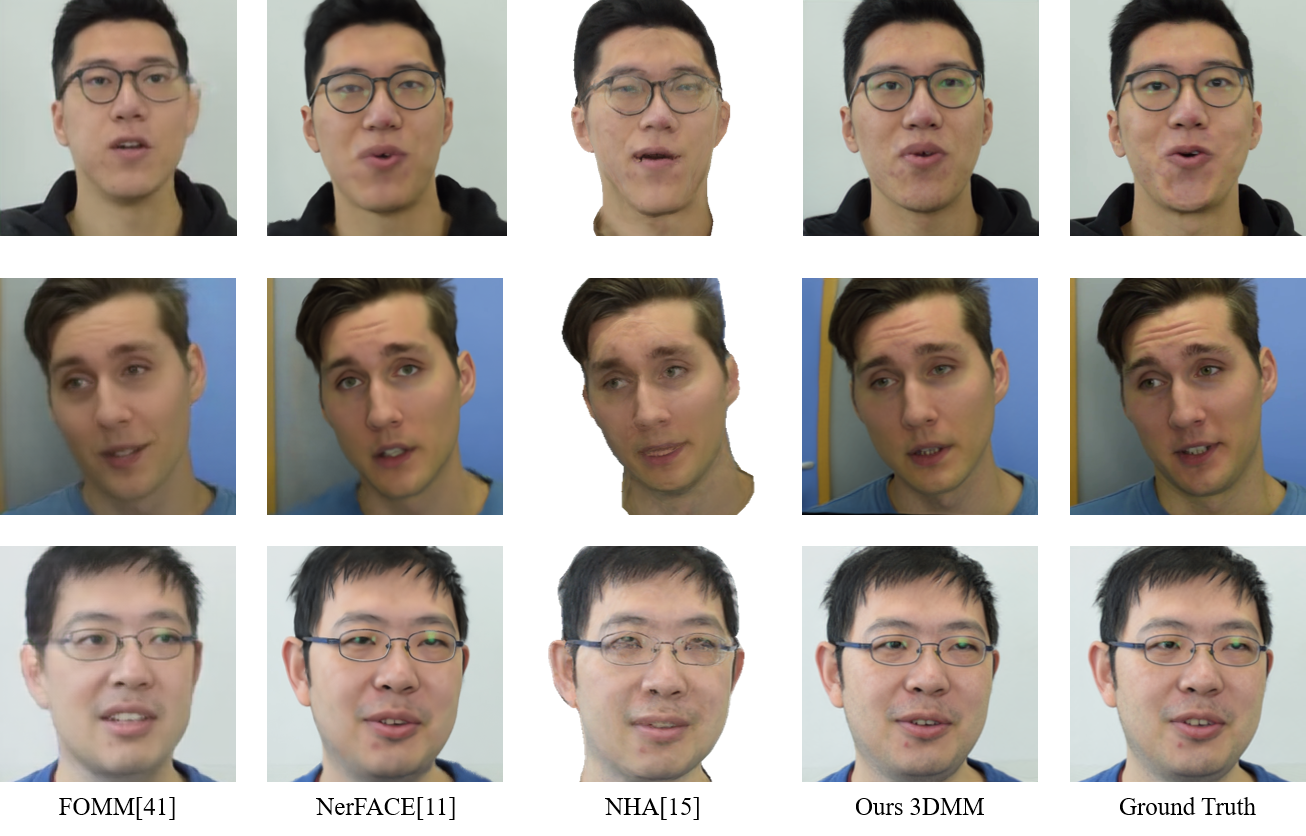}
    \caption{ Visualizations of 3DMM-driven face reenactment under the ground truth camera views.
    While the compared methods suffer from severe identity changes and distortions, our method obtains faithfully face reenactment performance.}
    \label{fig:com_3dmm}   
\end{figure*}

\begin{table}
    \small
    \centering 
    \caption{Quantitative evaluation of our method and PTI for 3D-aware face reconstruction.}
\begin{tabular}{c|c|c|c} 
\hline
\multirow{2}{*}{ Methods } & \multicolumn{3}{c}{ Metrics } \\
\cline { 2 - 4 } & PSNR $\uparrow$ & SSIM$\uparrow$ & LPIPS$\downarrow$ \\
\hline
PTI \cite{roich2022pivotal} & \makecell[c]{32.62}& \makecell[c]{0.959}& \makecell[c]{0.037} \\
Ours RGB & \makecell[c]{\textbf{34.70}}& \makecell[c]{\textbf{0.979}}& \makecell[c]{\textbf{0.024}} \\
\hline
\end{tabular}
    \label{tab:quantitative0}
\end{table}

\subsection{Results of RGB-based Face Reconstruction}
\label{sec:experiment-reconstruction}
We first conduct experiments with RGB frames as input. 
We adopt the three monocular videos used in NerFACE \cite{gafni2021dynamic} for evaluation.
For each video, we extract the first 2 minutes ($\sim$ 6000 frames) for training and the left 20 seconds ($\sim$ 1000 frames) for testing.
After training, we directly send each testing frame into the learned encoder to obtain its latent code, which is then sent to the generator for novel view synthesis.
To better verify the face reconstruction performance, we also present the performance of PTI \cite{roich2022pivotal}, which is an optimization-based GAN inversion method.
For each testing frame, PTI optimizes both the latent code and the EG3D generator for a faithfully reconstruction.

To measure their quality, we conduct a quantitative evaluation using several common metrics, including Peak Signal-to-Noise Ratio (PSNR), Structure Similarity Index (SSIM), and the Learned Perceptual Image Patch Similarity (LPIPS) \cite{zhang2018unreasonable}. 
As we don't have novel view ground truth images, we calculate these metrics under the same views of the testing frames.
{The numerical results are given in Table~\ref{tab:quantitative0}.}
Compared to PTI, we obtain superior performance under all three metrics.
In Figure~\ref{fig:pti}, we show some rendered images with the camera views of testing frames and two randomly picked novel views.
The complete novel view comparisons are given in the {\it supplementary materials}.
Our method can better maintain the personalized characteristics, such as the mouth area.
More importantly, for novel view synthesis, PTI shows significant artifacts for face shapes and cannot preserve the ID, while we obtain superior multi-view-consistent results.
These demonstrate that the learned personalized generative prior enables faithful face reconstruction.

\subsection{Results of 3DMM-driven Face Reenactment}
\label{sec:experiments-3dmm}

We then evaluate the performance of 3DMM-driven face reenactment.
We compare to the 3D-aware methods, Neural Head Avatar (NHA)~\cite{grassal2022neural} and NerFACE~\cite{gafni2021dynamic}. 
We also provide the results of FOMM \cite{siarohin2019first}, a 2D-based face animation method.
Note that FOMM doesn't support novel view synthesis.
As in the previous section, we adopt the three monocular videos used in NerFACE.
We extract the 3DMM expression coefficients from each frame for training and testing.

The average results in terms of PSNR, SSIM and LPIPS are listed in Table \ref{tab:quantitative}.
The 3D-aware methods NerFACE, NHA and ours obtain better performance than FOMM.
With the help of the high-quality prior of the generative model, our method significantly boosts the performance of 3DMM-driven face reenactment. 

We also show a qualitative comparison in Figure~\ref{fig:com_3dmm}, where all results are rendered under the same view of the ground truth image.
It can be seen that the results of FOMM have obvious artifacts, and the identity of the animated face is altered a lot from the ground truth. 
The results of NerFACE are too smooth, and the details of the facial textures are not well reconstructed. 
NHA cannot faithfully reconstruct facial characteristics, including eyeglasses and mouth areas.
In comparison, our method can better maintain the facial characteristics and generate faithful face reenactment.
{In Figure~\ref{fig:teaser}, we present some novel view rendered images of NerFACE and ours, a complete comparison is given in the {\it supplementary materials}.
The results of NerFACE are blurred and distorted.
It fails to generate high-quality renderings under novel views, while our method obtains faithfully multi-view-consistent images.
}





\begin{table}
    \small
    \centering 
    \caption{Quantitative evaluation of our method in comparison to 3DMM-driven face reenactment.}
\begin{tabular}{c|c|c|c} 
\hline
\multirow{2}{*}{ Methods } & \multicolumn{3}{c}{ Metrics } \\
\cline { 2 - 4 } & PSNR $\uparrow$ & SSIM$\uparrow$ & LPIPS$\downarrow$ \\
 \hline FOMM~\cite{siarohin2019first} & \makecell[c]{27.75}  & \makecell[c]{0.919}& \makecell[c]{0.059}\\
NerFACE~\cite{gafni2021dynamic} & \makecell[c]{29.76}& \makecell[c]{0.931}& \makecell[c]{0.053} \\
NHA~\cite{grassal2022neural} & \makecell[c]{31.52}& \makecell[c]{0.954}& \makecell[c]{0.039} \\
Ours 3DMM & \makecell[c]{\textbf{34.38}}& \makecell[c]{\textbf{0.972}}& \makecell[c]{\textbf{0.027}} \\
\hline
\end{tabular}
    \label{tab:quantitative}
\end{table}

\begin{figure*}[t]
    \centering
    \includegraphics[width=0.95\linewidth]{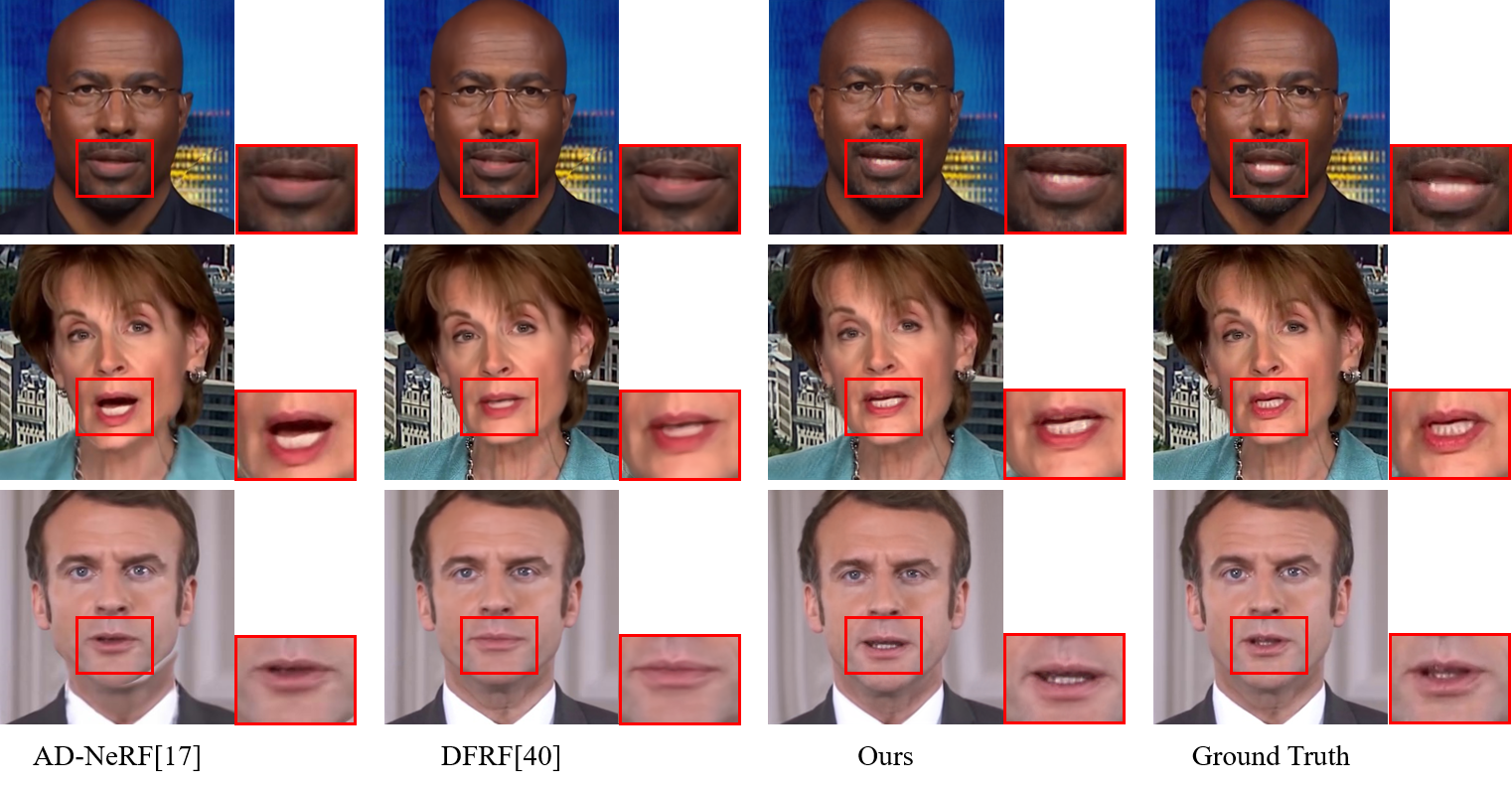}
    \caption{ Visualizations of audio-driven face reenactment under the ground-truth camera views.  The mouth areas are zoomed-in for better viewing. We obtain more faithfully rendering, especially for the shape and appearance in the mouth areas.}
    \label{fig:com_audio}   
\end{figure*}

    
\subsection{Results of Audio-driven Face Reenactment}
\label{sec:audio-driven experiments}
Following the practice of \cite{guo2021ad}, we perform audio-driven experiments on three public videos collected from YouTube.
The position of the camera is fixed and the resolution of the videos is $512 \times 512$. 
Each video is divided into two segments of the training set and testing set, with no overlap between them.
We extract their audio features for training and testing.

\begin{table}
    \small
    \centering 
    \caption{Quantitative evaluation of our method in comparison to audio-driven face reenactment.}
\begin{tabular}{c|c|c|c|c} 
\hline
\multirow{2}{*}{ Methods } & \multicolumn{4}{c}{ Metrics } \\
\cline { 2 - 5 } & PSNR $\uparrow$ & SSIM$\uparrow$ & LPIPS$\downarrow$ & SyncNet$\uparrow$ \\
 \hline 
 Ground Truth & \makecell[c]{-}  & \makecell[c]{-}& \makecell[c]{-} & \makecell[c]{7.653}
\\
 AD-NeRF~\cite{guo2021ad} & \makecell[c]{29.69}  & \makecell[c]{0.934}& \makecell[c]{0.057}& \makecell[c]{1.238}\\
DFRF~\cite{shen2022learning} & \makecell[c]{30.23}& \makecell[c]{0.939}& \makecell[c]{0.042} & \makecell[c]{4.142
}\\
Ours Audio & \makecell[c]{\textbf{32.57}}& \makecell[c]{\textbf{0.957}}& \makecell[c]{\textbf{0.035}} & \makecell[c]{\textbf{4.866
}}\\
\hline
\end{tabular}
    \label{tab:quantitative2}
\end{table}

\begin{figure*}[t]
    \centering
    \includegraphics[width=\linewidth]{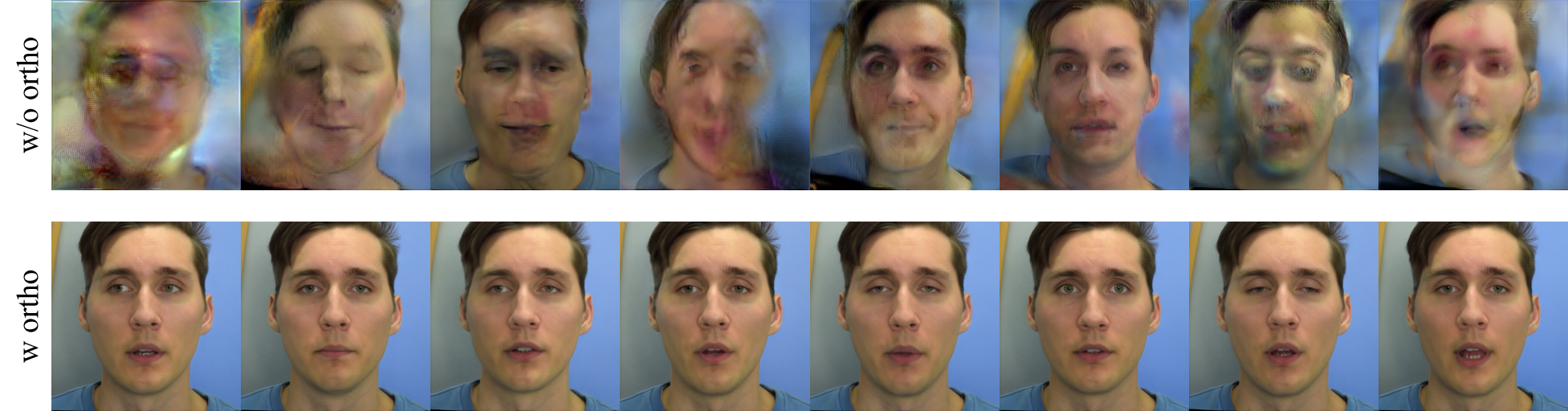}
    \caption{ Visualizations of the basis vectors learned with and without the orthogonal constraint. The orthogonal constraint can largely boost the disentanglement of the basis vectors.}
    \label{fig:ortho}   
\end{figure*}

\begin{figure*}[t]
    \centering
    \includegraphics[width=\linewidth]{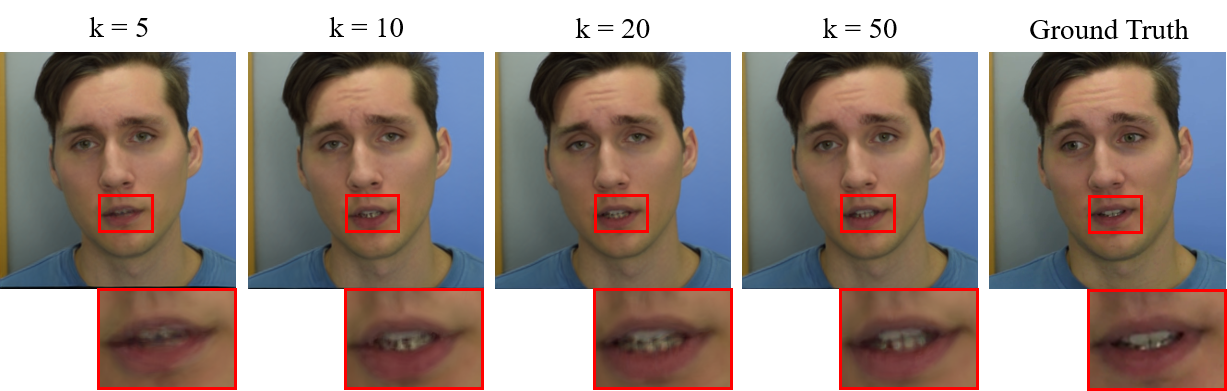}
    \caption{ Ablation study on the number of basis vectors. 
    The mouth areas are zoomed-in for better viewing. As the number of basis vectors increases, facial details such as teeth, wrinkles, can be better maintained.}
    \label{fig:k-number}   
\end{figure*}




We compare our method with two audio-driven NeRF methods: AD-NeRF~\cite{guo2021ad} and DFRF~\cite{shen2022learning}. 
DFRF is a few-shot method.
To make a fair comparison to it, we use a short 20 seconds video clip for training for all methods. 
In addition to the image quality metrics of PSNR, SSIM and LPIPS, 
SyncNet~\cite{chung2016out} is further used to measure audio-visual synchronization. 
The average metrics of three videos are given in Table \ref{tab:quantitative2}. 
In the audio-driven scenario, our method also outperforms the previous methods with a significant margin for all four evaluation metrics.
Besides, we provide some rendered images in Figure~\ref{fig:com_audio}.
We highlight the mouth areas that are most significant to audio-driven face reenactment.
Compared to AD-NeRF and DFRF, we obtain much better mouth shapes and teeth. 
We also provide some novel view results of DFRF and ours in Figure~\ref{fig:teaser}.
Under novel views, the distortion of DFRF is even worse.


\subsection{Ablation Studies}
The key to our approach is to learn a basis to represent the personalized generative prior.
We conduct ablation studies based on RGB-based face reconstruction to analyze the properties of the basis vectors.

\textbf{Visualization of the basis vectors.}
We require the set of basis vectors to be orthogonal to each other during training. 
In Figure \ref{fig:ortho} we make a visualization of the learned basis.
We also show the visualization results for a basis learned without the orthogonal constraints.
It can be seen that the basis vectors are coupled together when there is no orthogonal constraint. 
With the orthogonal constraint, the basis vectors are disentangled and show better semantic meanings.
The quantitative comparison results are shown in Table \ref{tab:quantitative3}, which demonstrates that the reconstruction quality is much better with the orthogonal constraint.


\textbf{Number of the basis vectors.}
We further explore the effect of the number of basis vectors. 
We vary the number of basis vectors by 5, 10, 20 and 50.
The quantitative comparisons in terms of PSNR, SSIM and LPIPS are given in Table~\ref{tab:quantitative3}.
And Figure~\ref{fig:k-number} shows some rendered images under the different value of $k$.
We highlight the mouth areas for better visualization.
As the number of basis vectors increases, the model is more capable of representing facial details and obtains better rendered quality.

\begin{table}
    \small
    \centering 
    \caption{Quantitative comparison of the ablation study on the orthogonal constraint and the number of basis vectors.}
\begin{tabular}{c|c|c|c} 
\hline
\multirow{2}{*}{ Schemes } & \multicolumn{3}{c}{ Metrics } \\
\cline { 2 - 4 } & PSNR $\uparrow$ & SSIM$\uparrow$ & LPIPS$\downarrow$ \\
 \hline k = 50 (w/o ortho)  & \makecell[c]{33.36}& \makecell[c]{0.962}& \makecell[c]{0.037} \\
 k = 5 (w ortho) & \makecell[c]{28.64}  & \makecell[c]{0.927}& \makecell[c]{0.055}\\
k = 10 (w ortho) & \makecell[c]{30.83}& \makecell[c]{0.946}& \makecell[c]{0.042} \\
k = 20 (w ortho) & \makecell[c]{33.98}& \makecell[c]{0.965}& \makecell[c]{0.033} \\
k = 50 (w ortho) & \makecell[c]{\textbf{34.70}}& \makecell[c]{\textbf{0.979}}& \makecell[c]{\textbf{0.024}} \\
\hline
\end{tabular}
    \label{tab:quantitative3}
\end{table}


\paragraph{Supplementary materials.}
Due to space limitation, we only provide some rendered images and novel views in the main paper.
We highly recommend checking the {\it supplementary materials} for more novel view synthesis results.

\section{Conclusions}
\label{sec:conclusions}
In this work, we propose to utilize 3D-aware generative prior for facial avatar reconstruction and reenactment from monocular videos.
We propose an efficient method to learn a local and low-dimensional subspace in the latent space of 3D-GAN, for a better maintenance of personalized characteristics.
The learned personalized generative prior provides a good constraint for 3D-aware generation, which is helpful for modeling the complex facial dynamics and missing 3D information in monocular videos.
We conduct extensive experiments, including RGB-based face reconstruction and 3DMM and audio-driven face reenactment.
Compared to existing works, we obtain superior performance both quantitatively and qualitatively.

\textbf{Discussions.} 
In the current formulation, we aim to optimize a personalized subspace that can be used for faithful reconstruction and self-reenactment driven by different signals.
The learned personalized basis has shown some good properties of disentanglement and semantic meanings.
We will continue our study on the properties of 3D-aware personalized generative prior and investigate more strategies to control the basis vectors.
Besides, it is also very interesting and helpful to explore cross-identity face reenactment based on personalized 3D-aware generative prior.

{\small
\bibliographystyle{ieee_fullname}
\bibliography{egbib}
}

\end{document}